\documentclass[10pt,journal,compsoc]{IEEEtran}
\usepackage[T1]{fontenc}
\usepackage[latin9]{inputenc}
\usepackage[english]{babel}
\usepackage{array}
\usepackage{amsmath}
\usepackage{amssymb}
\usepackage{graphicx}
\usepackage{nomencl}
\usepackage{xcolor}
\usepackage{wrapfig}
\usepackage{arydshln}
\usepackage[nocompress]{cite}

\providecommand{\makenomenclature}{\makeglossary}
\makenomenclature
\usepackage[unicode=true]
 {hyperref}
\usepackage{breakurl}

\makeatletter
\usepackage[english]{babel}
\AtBeginDocument{

}

\makeatother
\def \fig_width {0.75}
\begin{document}

\title{QANet - Quality Assurance Network for Image Segmentation}

\author{Assaf Arbelle, Eliav Elul and Tammy Riklin~Raviv
\IEEEcompsocitemizethanks{\IEEEcompsocthanksitem Assaf Arbelle, Eliav Elul and Tammy Riklin~Raviv are with the School of Electrical and Computer Engineering and the Zlotowski Center for Neuroscience, Ben Gurion University of the Negev, Israel.\protect\\}%
}
\IEEEpeerreviewmaketitle
\IEEEtitleabstractindextext{
\begin{abstract}
We introduce a novel Deep Learning framework, which quantitatively estimates image segmentation quality  without the need for human inspection or labeling.  We refer to this method as a Quality Assurance Network - QANet. 
Specifically, given an image and a `proposed' corresponding segmentation, obtained by any method including manual annotation, the QANet solves a regression problem in order to estimate a predefined quality measure with respect to the unknown ground truth. 
The QANet is by no means yet another segmentation method. Instead, it performs a multi-level, multi-feature comparison of an image-segmentation pair based on a unique network architecture, called the RibCage.

To demonstrate the strength of the QANet, we addressed the evaluation of instance segmentation using two different datasets from  different domains, namely, high throughput live cell microscopy images from the Cell Segmentation Benchmark and natural images of plants from the Leaf Segmentation Challenge.
While synthesized segmentations were used to train the QANet, it was tested on segmentations obtained by publicly available methods that participated in the different challenges. 
We show that the QANet accurately estimates the scores of the evaluated segmentations with respect to the hidden ground truth, as published by the challenges' organizers.

The code is available at: TBD.
\end{abstract}
\begin{IEEEkeywords}
Deep Learning, Leaf Segmentation, Microscopy, Segmentation, Quality Assurance
\end{IEEEkeywords}
}

\maketitle
\IEEEdisplaynontitleabstractindextext
\IEEEpeerreviewmaketitle

\setlength{\intextsep}{1mm}
\section{Introduction}\label{sec:Inrtoduction}
\begin{figure}
\includegraphics[trim={0bp 0bp 0bp 62bp},clip, width=0.48\columnwidth]{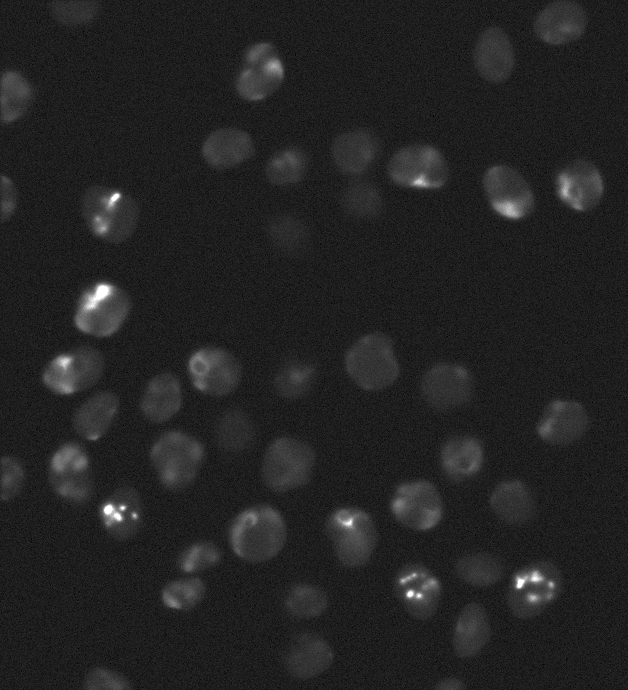}
\includegraphics[width=0.48\columnwidth]{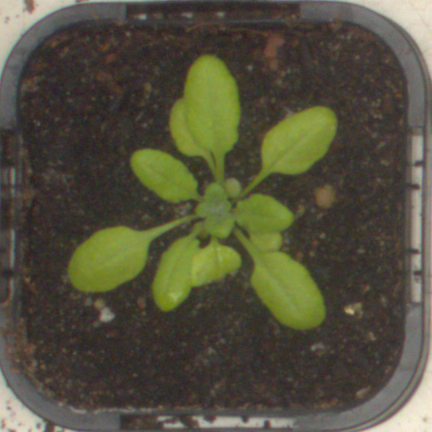}
\caption{\label{fig:Datasets} Examples of raw images from the Cell Segmentation Benchmark$^1$ (left) and the Leaf Segmentation Challenge$^2$ (right)} 
\end{figure}

Image segmentation is a well-studied problem, playing a major
role in almost any image analysis task. Model-based as well as data-driven approaches are usually validated on `previously unseen' annotated test sets, being compared to current state-of-the-art segmentation methods adapted to the task examined, the imaging modality and depicted scene or objects. Some recent methods manage to achieve almost `human-level' scores on well known challenges and benchmarks. Nevertheless, a measurable and objective evaluation of image segmentation with respect to user specific data has yet to be addressed.  

Consider, for example, the analysis of live cell microscopy images (Figure~\ref{fig:Datasets} left). Instance segmentation of individual cells allows the extraction of
useful information at the cellular level. Further analysis of this output may shed light on biological
processes and phenomena - potentially making a significant impact on
health-care research. The implications of some of these biological
findings are critical, thus every step in the research process, in particular cell segmentation, must be reliable. 
Online benchmarks evaluate, compare and objectively rank multiple automatic methods.
Still, the question arises whether one can be assured that a given segmentation method will consistently perform well enough on private data.
Moreover, current methods are ranked based on the statistics of the measured score on some specific data set, how could a user detect specific cases of segmentation failures which may risk the overall analysis?
In other words, can we avoid visual inspection of the
results or an additional test with data-specific manual annotations,
in this, and alike, important pipelines? 

This paper is the first to address (to the best of our knowledge)
quantitative assessment of image segmentation {\em without} ground truth
annotations given only the image and its corresponding evaluated segmentation, provided by any source. Specifically, we introduce a deep neural network, termed Quality Assurance Network - QANet, that is able to estimate a predefined quality measure, e.g. the intersection-over-union (IoU) or Dice scores of
image-segmentation pairs. It should be stressed that QANet {\em does not}  aim to
estimate the  ground truth (GT)  segmentation and compare it with the
segmentation to be evaluated. Instead, it solves a regression problem, providing a quality assessment score of the evaluated segmentation. 

\begin{figure}
\includegraphics[width=0.48 \columnwidth]{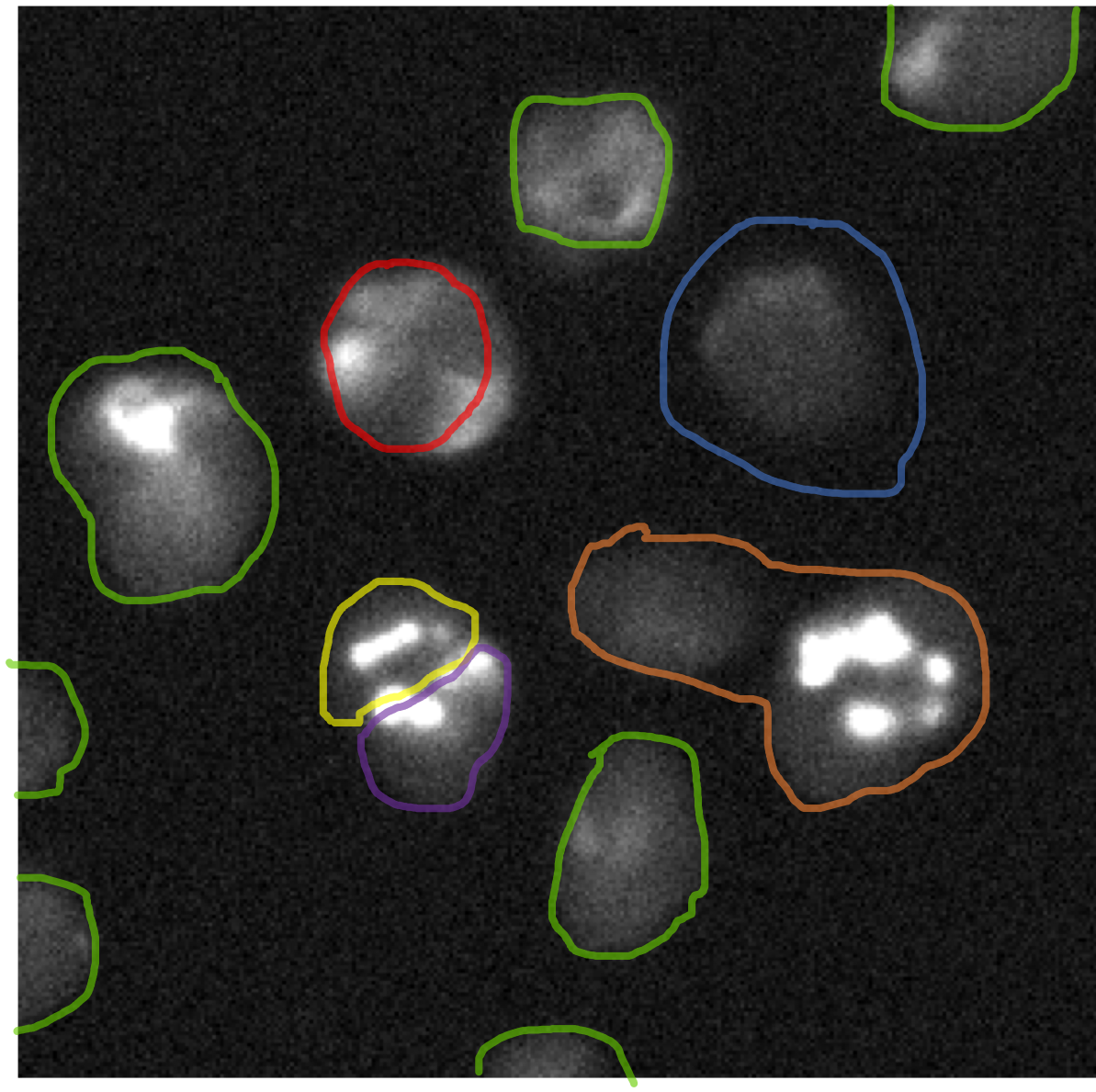} 
\includegraphics[width=0.48 \columnwidth]{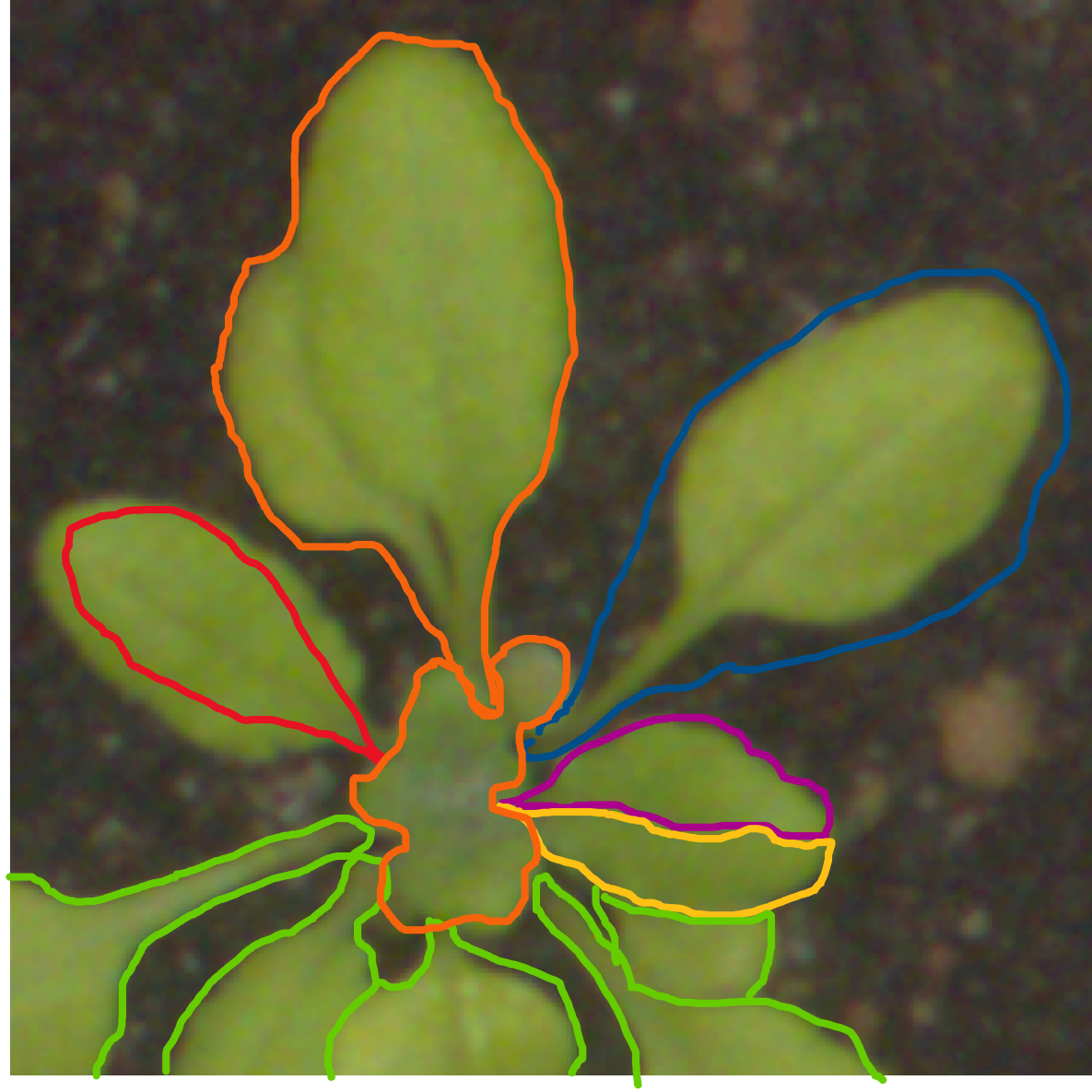}
\caption{\label{fig:overlap}Simulated Training Data: An example of  a microscopy image (left) and a plant image (right) with the contours of some segmentation method. The segmentation errors are color-coded to depict different types of possible errors: Green - accurate segmentation; Red - under detection, parts of the object are not included in the segmentation; Blue - over detection, the segmentation includes background areas; Orange - merged instances, two (or more) objects are segmented as one object; Yellow/Purple - object split, one object is segmented as two separate objects.}
\end{figure}

The QANet is inspired by the ability of a human expert to assess the quality of a given segmentation by comparing it to the raw image alone (Figure~\ref{fig:overlap}).
In order to facilitate the comparison capability, it is based on a unique architecture, called the RibCage Network,
which was first introduced in~\cite{arbelleISBI2018} as a discriminator in an adversarial setting. 
The RibCage is designed as a three channel network - two `ribs' and a `spine'. Each rib gets as input either the image or the evaluated segmentation, and the spine merges the two channels at each layer allowing a multi-level, spatially sensitive comparison of
the inputs. 
We show that this architecture has an advantage over a simple concatenation of the image and its segmentation, which is limited to low level feature comparison. It also outperforms the Siamese architecture in which each input is processed independently and is therefore restricted to the comparison of high level features.

In this work, we specifically address quantitative evaluation of instance segmentation. As opposed to semantic segmentation, we need to consider the partition of an image into an unknown number of similar, possibly overlapping objects with arbitrary labeling. In order to preserve the separation of neighboring instances, we choose to define the segmentation as a trinary image representing foreground, background and instance boundary.

To demonstrate the strength of the QANet, we addressed the evaluation of instance segmentation using two different datasets from two different domains, namely, high throughput live cell fluorescence microscopy images from the Cell Segmentation Benchmark\footnote{\url{http://celltrackingchallenge.net/latest-csb-results/}\label{url:ctc_web}}~\cite{Ulman17} and natural (RGB) images of plants from the Leaf Segmentation Challenge\footnote{\url{https://www.plant-phenotyping.org/CVPPP2017}\label{url:lsc_web}}~\cite{bell2016aberystwyth,MinerviniPRL2015,scharr2014annotated}.
Given a specific domain, we train the QANet by manipulating the GT segmentations of the training set, covering the entire range of the desired quality measure.
While synthesized segmentations were used to train the QANet, it was tested on segmentations obtained by publicly available methods that participated in the relevant challenge.  
We show that the QANet accurately estimates the scores of the evaluated segmentations with respect to the hidden GT (of the test set), as published by the challenge organizers. Promising results with maximum relative error of~2\% were obtained for both datasets.

%

The rest of the paper is organized as follows:
Section~\ref{sec:PrevWork} discusses some related works and highlights the novelty of the task we aim to address. In Section~\ref{sec:Method} we formulate the QA problem and quality measure;
present the RibCage architecture as well as the loss; and discuss the
simulation of training examples. Experimental results are presented in
Section~\ref{sec:Experiments}. Specifically we present results on simulated as well as real data from two domains. We further perform an ablation study testing the RibCage with respect to two alternative architectures and binary vs. trinary segmentation representations. We conclude in
Section~\ref{sec:Summary}.

\section{Relation to Previous Work}\label{sec:PrevWork}
Some Computer Vision methods, not necessarily for image segmentation,
produce confidence scores alongside their output.  One such example is the
YOLO network, which is designed  for object detection~\cite{Redmon2016YOLO,Redmon2017yolo9000}. The YOLO predicts both a bounding box and
its estimated IoU. Nevertheless, there are essential differences
between the QANet and the YOLO confidence score. Not only is a
bounding-box a very crude estimation of the instance segmentation, but
while the YOLO's scoring can be applied only to its own outputs the QANet works on the output of other methods (possibly even the YOLO's outputs) and scores the output independently of the method creating it. 

Similar to the YOLO, which provides its own confidence score, some  segmentation methods produce an uncertainty map or segmentation error margin along with the estimated object boundaries. This is obtained by introducing some stochastic to the segmentation process. 
A classical example is the work of Fan et al.~\cite{fan2007mcmc} which uses Markov Chain Monte Carlo (MCMC) to sample segmentations. 
In~ \cite{goldberg2018sampling} an MCMC method was used to generate multiple segmentation estimates, which together defined segmentation error margins. Segmentation uncertainty map were defined based on stochastic active contour in~\cite{hershkovitch2018model}. Recently, following the Monte Carlo Dropout of ~\cite{gal2016dropout} 
segmentation uncertainty had been incorporated in several deep learning frameworks~ \cite{bragman2018uncertainty, eaton2018towards, nair2019exploring}.   It should be stressed, however, that all of these methods output segmentations, and the corresponding uncertainty maps are exclusively evaluated with respect to the predicted outputs. Therefore, the resulting uncertainty maps are not meaningful for evaluating segmentations of other sources. 
Furthermore, although \cite{hershkovitch2018model} show a correlation between the desired quality measure (Dice coefficient in their case) to the level of uncertainty, they do not claim to predict the quality value itself. 

When considering networks which score the outputs of other networks, the discriminator
in an adversarial framework comes to mind
\cite{goodfellow2014}. However, we should note that the goal of the
discriminator is not to regress the confidence score, but only to
preform as an implicit loss function for its adversarial network. It is usually trained for binary classification of "real" versus "fake" examples, and once the training is complete, the discriminator collapses and does not produce an informative output.

\section{Method}\label{sec:Method}

\subsection{Formulation}\label{subsec:Formulation}
Our objective is to evaluate multiple instance object segmentation.
Let $I \colon \Omega \to \mathbb{R}$ be an image, where $\Omega$ is either a 2D or a 3D image domain. Let $\Gamma_{\scriptsize{\mbox{GT}}} \colon \Omega \to \mathbb{L}$ be the corresponding 
GT segmentation with labels $\mathbb{L}=\{0,1,2\}$ corresponding
to the background, foreground and object boundaries, respectively.
 Each connected component of the foreground represents a single instance. We wish to evaluate a segmentation,  $\Gamma_{\scriptsize{\mbox{E}}} \colon \Omega \to \mathbb{L}$, of $I$. The evaluation criteria can be any segmentation quality measure, i.e., Dice measure \cite{dice1945measures}, IoU \cite{jaccard1901etude}, Hausdorff distance \cite{Dubuisson94}, etc. 
We denote the true quality measure for a pair of GT and evaluated segmentation as $Q(\Gamma_{\scriptsize{\mbox{GT}}},\Gamma_{\scriptsize{\mbox{E}}})$. Our goal is to estimate the quality measure given only the raw image $I$ and the evaluated segmentation $\Gamma_{\scriptsize{\mbox{E}}}$ denoted as $\widehat{Q}(I,\Gamma_{\scriptsize{\mbox{E}}})$. 

\subsection{RibCage Network Architecture and Loss}\label{subsec:RibCage Network Architecture}
\begin{figure}
\centering
\includegraphics[width=\fig_width \columnwidth]{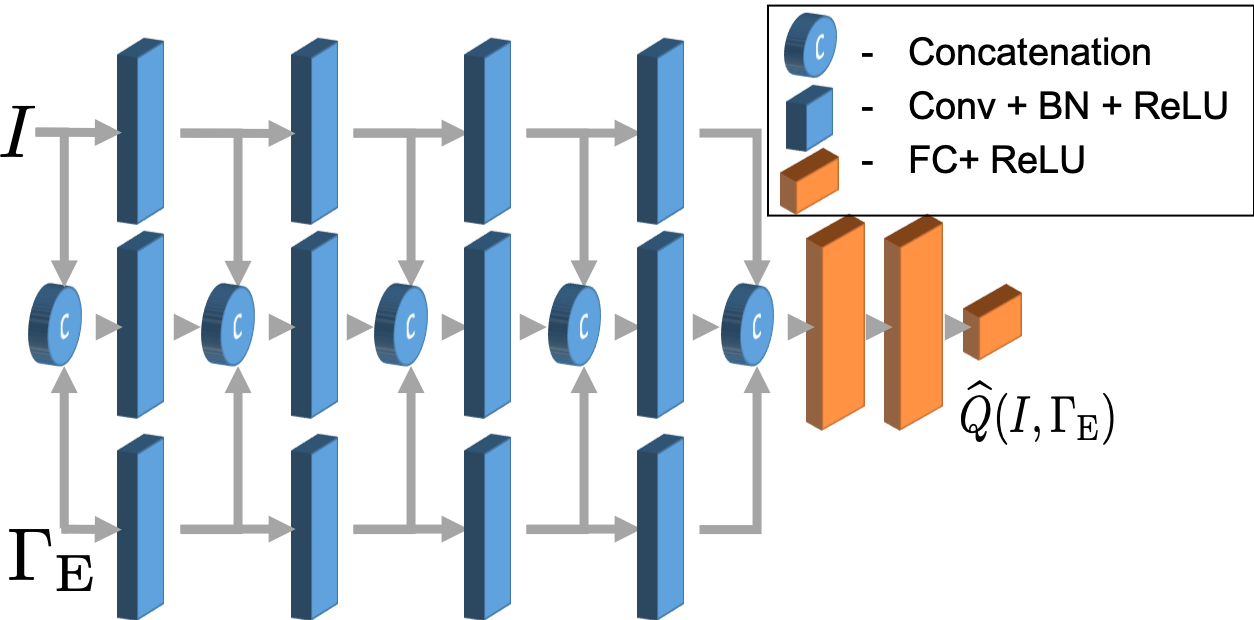}
\caption{\label{fig:RibCageArchitecture}\textbf{Network Architecture:} The QANet is designed in the form of a RibCage Network, the top rib processes the raw image while the bottom rib processes the segmentation proposal. The spine processes the concatenation of the two ribs throughout the depth of the network. The four RibCage blocks are followed by three FC layers which output a single scalar representing the estimated $\widehat{Q}_{\theta}(I,\Gamma_{\scriptsize{\mbox{E}}})$ measure.} 
\end{figure}
Let  $\widehat{Q}_\theta(I,\Gamma_{\scriptsize{\mbox{E}}})$ denote the output of a QANet with parameters $\theta$. The QANet is implemented using the RibCage architecture~\cite{arbelleISBI2018}. The strength of this  architecture is its ability  to extract and compare multilevel features from two inputs in a spatially sensitive manner over multiple scales. The RibCage architecture, outlined in Figure~\ref{fig:RibCageArchitecture}, is comprised of blocks of two ribs connected to a spine. The spine merges the two ribs' outputs. Each of the three channels is  passed through a strided convolutional layer, batch normalization and a ReLU activation.

Let $l\in [1,L]$ denote the index of a RibCage block where $L$ is the total number of blocks. 
 The inputs $r^{l-1}_{1}$, $r^{l-1}_{2}$ and $s^{l-1},$  which represent the two ribs and the spine respectively, are defined as follows:
\begin{equation}
r^{l}_{1} = f(\theta_{r^{l}_{1}} \ast r^{l-1}_{1})
\end{equation}
\begin{equation}
r^{l}_{2} = f(\theta_{r^{l}_{2}} \ast r^{l-1}_{2})
\end{equation}
\begin{equation}
s^{l} = f(\theta_s \ast s^{l-1} + \theta_{s1} \ast r^{l-1}_{1} + \theta_{s2} \ast r^{l-1}_{2})
\end{equation}
where the function $f(\cdot)$ represents the Batch Normalization and the ReLU activation. The initial inputs $r^{0}_{1}$, $r^{0}_{2}$ are set to be the image $I$ and the segmentation $\Gamma_{\scriptsize{\mbox{E}}}$. The spine input is set to be $s^{0}=0$. The outputs of the last block $L$ are then passed to several FC layers resulting in a single scalar $ \widehat{Q}_\theta$.

The QANet is trained to regress the values of some predefined quality measure, $Q$. The training loss, $\mathcal{L}$, is the mean squared error (MSE) between the networks output and the true measure:
\begin{equation}
\mathcal{L} = ||\widehat{Q}_{\theta}(I,\Gamma_{\scriptsize{\mbox{E}}})-Q(\Gamma_{\scriptsize{\mbox{GT}}},\Gamma_{\scriptsize{\mbox{E}}})||_2
\end{equation}

\subsection{Synthesized Segmentations}\label{subsec:Synthesized Segmentations}
The input to the QANet, in both training and test phases, is composed of image-segmentation pairs. In the training phase, we synthesize  imperfect segmentation proposals by deforming the given GT segmentations such that the true value of the quality measure, $Q$, can be calculated. 
We wish that the deformations will be as realistic as possible, as if they were the output of some segmentation algorithm or an unqualified annotator. The deformation process consists of two stages, morphological operations (MO) and non-rigid perturbations. 
\subsubsection{Morphological Operations}
The morphological operations, consisting of erosion, dilation, opening and closing, simulate under/over-estimation of the object region. We randomly sample a five-state variable, $OP$, to decide on either one of the morphological operations or the identity. If one of the morphological operations is selected, we sample a positive integer, $\sigma_{\scriptsize{\mbox{MO}}}$, to define the MO kernel size. We note that after the MO stage an instance may be completely removed or merged with neighboring instances.
\subsubsection{Non-Rigid Perturbations}
To further enrich the variability in segmentation errors, we perform non rigid perturbations. As in~\cite{Ronneberger15} we randomly sample a vector field map, $[v_x, v_y]$, in the size of the input domain $\Omega$ and smooth it using a Gaussian kernel, $\sigma_g$.
The final deformed segmentation $\Gamma_{\scriptsize{\mbox{E}}}$ is obtained by applying the smoothed vector field map to the segmentation resulting from the MO stage.

\section{Experiments}\label{sec:Experiments}
We evaluated the QANet method using two datasets from different domains, each with its own definition of segmentation quality measures, $Q(\Gamma_{\scriptsize{\mbox{GT}}},\Gamma_{\scriptsize{\mbox{E}}})$. Specifically we used live cell microscopy data from the Cell Segmentation Benchmark and plant images from the Leaf segmentation Challenge. Figure~\ref{fig:Datasets} presents example images from each of the datasets. 
In Section~\ref{subsec:QANetEvaluation} we present the evaluation methodology for the QANet and the measures that quantify the network's accuracy, which are used throughout the experiments. 
An ablation study is presented in Section~\ref{subsec:AblationStudy} to justify the configuration of the proposed method and evaluate its components. Specifically we tested the RibCage network architecture selection and the input segmentation representation.
We show the prediction estimation capabilities of the QANet on the Cell Segmentation Benchmark in Section~\ref{subsec:CSB Leaderboard} and the Leaf Segmentation Challenge in Section~\ref{subsec:LSC}.

\subsection{Evaluation of the QANet}\label{subsec:QANetEvaluation}
The MSE between the GT quality value, $Q$, and the predicted quality value $\widehat{Q}_\theta$ is the most straight
forward measure for evaluating the QANet. However, for demonstrating
the QANet performances for different $Q$ values we used a scatter plot
showing $Q(\Gamma_{\scriptsize{\mbox{GT}}},\Gamma_{\scriptsize{\mbox{E}}})$, with respect to the predicted $\widehat{Q}_{\theta}(I,\Gamma_{\scriptsize{\mbox{E}}})$, for the test
examples. In addition, we calculated the Hit Rate - which is the
normalized number of test examples of which the differences between $Q$ and $\widehat{Q}_\theta$ are within a specified tolerance. We also
used the Area Under the Curve (AUC) of the Hit Rate versus tolerance plot as another quantitative measure of the evaluation of the QANet. 
\subsection{Ablation Study}\label{subsec:AblationStudy}
We examined two main aspects of the QANet: the network architecture, and the segmentation representation. The experiments were conducted using the Cell Segmentation Benchmark along with the benchmark's SEG measure as the target quality measure. 
\subsubsection{The SEG Measure}\label{subsec:The SEG Measure}
The SEG measure, introduced in~\cite{Mavska14} for microscopy cell segmentation evaluation, is an extension of the IoU measure for multiple instances.
The measure first finds a one-to-one matching between the evaluated object labels and the GT labels, where each matched pair is scored using the IoU measure.
Let $K$ and $K'$ be the number of individual cells in the GT and
evaluated segmentation, $\Gamma_{\scriptsize{\mbox{GT}}}$ and $\Gamma_{\scriptsize{\mbox{E}}}$ respectively.
Let $c\in\Gamma_{\scriptsize{\mbox{GT}}}$ and $c'\in\Gamma_{\scriptsize{\mbox{E}}}$ define corresponding connected components in
the evaluated and GT segmentations, respectively. 
The SEG measure is defined as the IoU of
the GT and the evaluated objects, unless their overlap is lower than
50\%.  The mean SEG measure over all GT objects is formulated as follows:
\begin{equation}
Q(\Gamma_{\scriptsize{\mbox{GT}}},\Gamma_{\scriptsize{\mbox{E}}}) =\frac{1}{K}
\sum_{c\in\Gamma_{\scriptsize{\mbox{GT}}}}\sum_{c'\in\Gamma_{\scriptsize{\mbox{E}}}} \begin{cases} IoU(c,c') &
  \alpha(c,c') >0.5 \\
0 & \mbox{otherwise}
\end{cases}
\end{equation}
where, $IoU(c,c')=\frac{|c\cap c'|}{|c\cup c'|}$  and $\alpha(c,c') =\frac{|c\cap c'|}{|c|}.$
We note that for every connected component in $c\in\Gamma_{\scriptsize{\mbox{GT}}}$ there exists at most one
connected component $c'\in\Gamma_{\scriptsize{\mbox{E}}}$ with overlap grater than 50\%.  If there is no such 
object, the cell is considered undetected and its SEG score is set to zero. 



\subsubsection{Microscopy Cell Segmentation Training Data}\label{sub:Training Data}
The input to the QANet, in both training and test phases, is composed of image-segmentation pairs. In the training phase we use simulated images such that the corresponding GT segmentations are known. To synthesize  imperfect segmentation proposals we deform the GT segmentations such that the true value of the quality measure, $Q$, can be calculated.
The dataset was split into a training and validation set, 70\%-30\% respectively.\\
\textbf{Images and GT Segmentations:}
The training images and GT segmentations were synthesized using the CytoPacq web service\footnote{\url{https://cbia.fi.muni.cz/simulator}}~\cite{wiesner2019cytopacq}. We tuned the system to produce data similar to  the Fluo-N2DH-SIM+ from the Cell Segmentation Benchmark. The CytoPacq synthesizes the images in three steps: 1) 3D digital phantom simulation that generates spatial
objects of interest and their structure. This step defines the GT segmentation $\Gamma_{\scriptsize{\mbox{GT}}}$; 2) Simulation that
models image formation in the optical system; 3) Acquisition device simulation that mimics image capturing process when using digital image detectors.
We generated 10000 images of size $420\times 420$ pixels, each image containing between $1$ to $60$ cells.\\
\textbf{Synthesizing Imperfect Segmentations}
If the five state variable $OP$ is not the identity, a random integer is sampled form a integer uniform distribution: $\sigma_{\scriptsize{\mbox{MO}}}\sim U(1,4)$ which determines the size of the MO kernel. The non-rigid perturbation is defined by a vector field sampled from a uniform distribution $v_x,v_y\sim U(-512,512)$ followed by a smoothing with a Gaussian kernel of width $\sigma_g = 38$.

\subsubsection{Microscopy Cell Segmentation Test Data}\label{sub:TestData}
Although the QANet was trained on synthesized images (Section~\ref{sub:Training Data}) and simulated segmentations, we tested the method's true capabilities on real data.\\
\textbf{Test Images:}
In addition to the simulated data detailed in \ref{sub:Training Data}, we tested our framework with two \textit{real} fluorescent microscopy datasets from the Cell Segmentation Benchmark, namely Fluo-N2DH-GOWT1 and
Fluo-N2DL-HeLa. The Fluo-N2DH-GOWT1 dataset is a data of GFP-GOWT1 mouse stem cells acquired by Leica TCS SP5 microscope with pixel size $0.24 \times 0.24$ microns~\cite{bartova2011recruitment}.
The Fluo-N2DL-HeLa dataset is of HeLa cells stably expressing H2b-GFP  acquired by  Olympus IX81 microscope with pixel size $0.645 \times 0.645$ microns~\cite{neumann2010phenotypic}.\\
\textbf{Test Segmentations:}
The test segmentations were generated using three
state-of-the-art cell segmentation methods applied to the test images. These include two Deep Learning based methods CVUT-CZ~\cite{Sixta2019Thesis} and BGU-IL(3)~\cite{arbelleIsbi2019}, and a classical method KTH-SE(1)~\cite{Magnusson16Thesis}. All methods were downloaded from the Cell Segmentation Benchmark website. In the following experiments we refer to the four method-dataset combinations as: KTH-GOWT1 and KTH-HeLa - result obtained by running the KTH-SE(1) method on Fluo-N2DH-GOWT1 and Fluo-N2DL-HeLa datasets respectively; CVUT-GOWT1 and CVUT-HeLa - result obtained by running the CVUT-CZ method on Fluo-N2DH-GOWT1 and Fluo-N2DL-HeLa datasets respectively.

\begin{figure*}[t!]
\begin{center}

\setlength{\tabcolsep}{0mm}
\renewcommand{\arraystretch}{0.2}
\includegraphics[width=\fig_width \columnwidth]{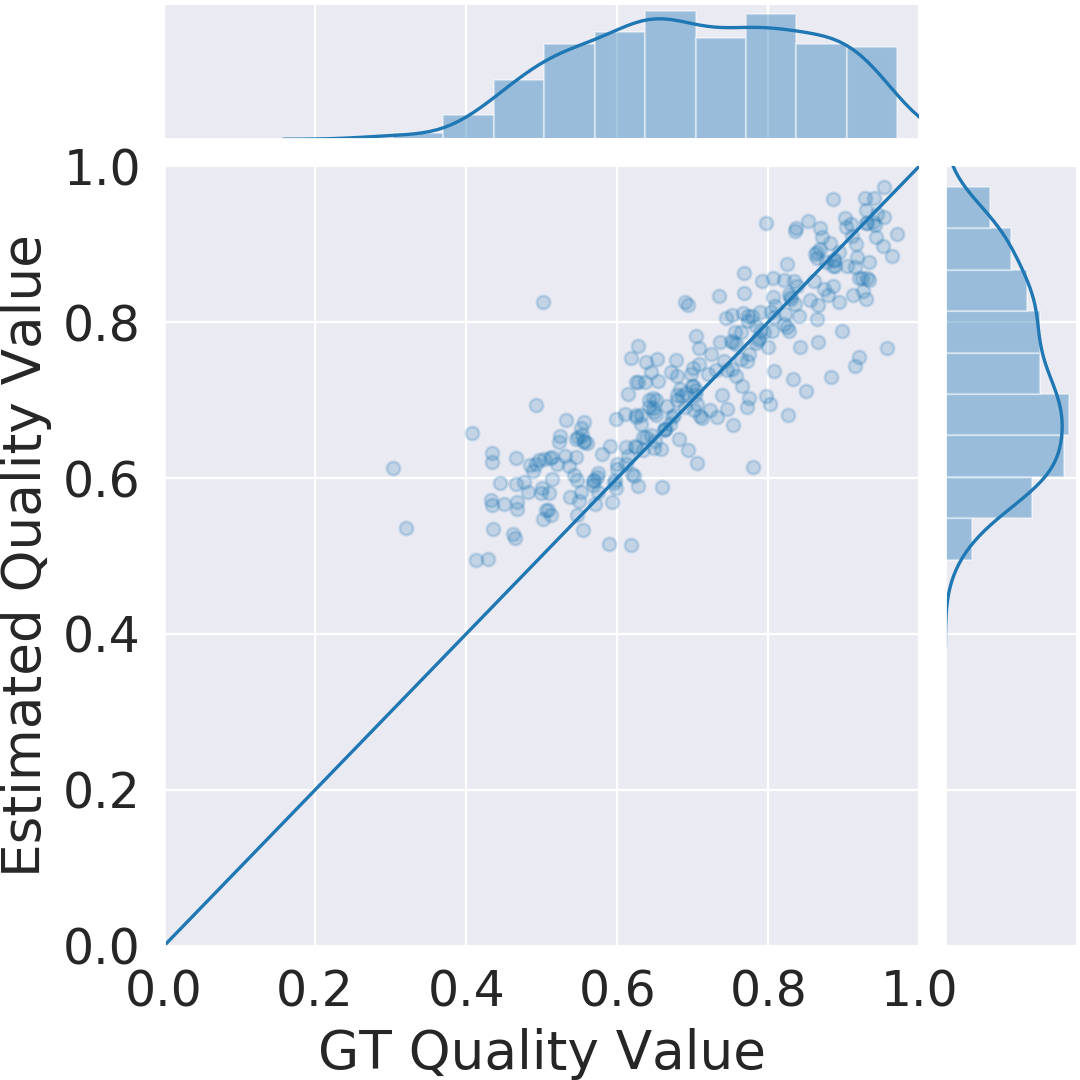}~~~~~~~~
\includegraphics[width=\fig_width \columnwidth]{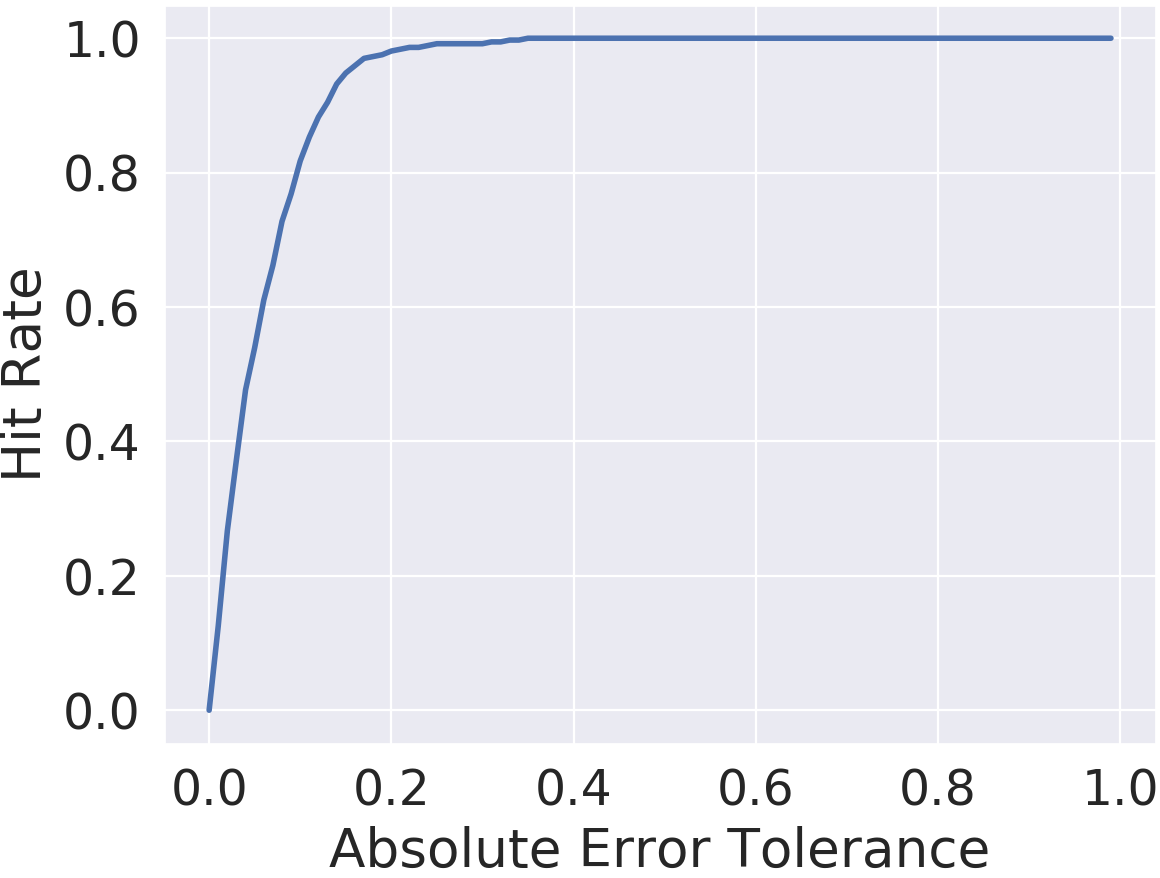}
\end{center}
\caption{\textbf{Cell Segmentation Validation Set:} The left image shows the scatter plot of the GT quality values with respect to the estimated quality value for the cell segmentation validation images. The horizontal axis is the GT SEG measure of the instance and the vertical axis show the QANet output. The diagonal line represents the optimal, desired, output. On the right is a Hit Rate curve as a function of SEG measure prediction tolerance.\label{fig:plot_val_test}}
\end{figure*}

\subsubsection{Network Architecture Experiment}\label{subsec:Network Architecutre}

\begin{figure}[t!]
\begin{center}
\includegraphics[trim={0bp 0bp 0bp 0bp},clip, width=\fig_width \columnwidth]{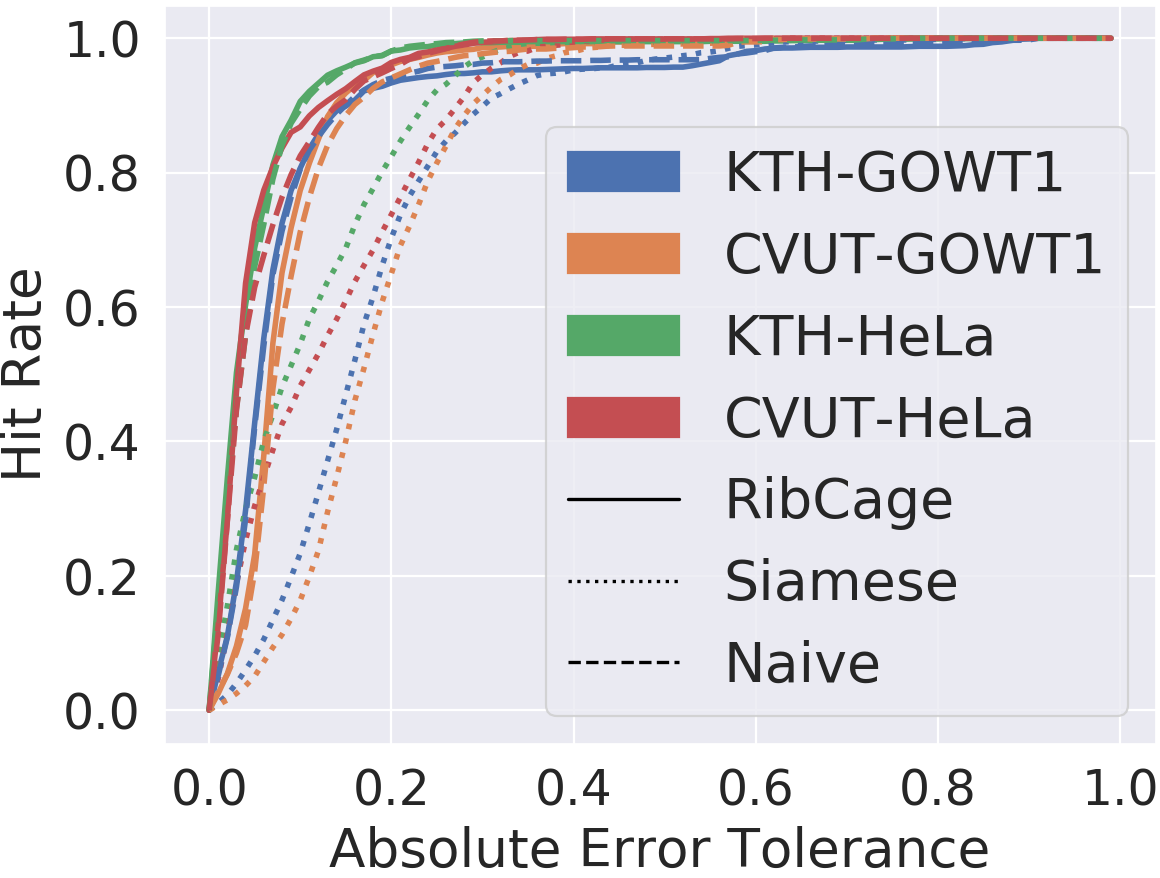}
\end{center}
\caption{The Hit Rate curves as a function of SEG prediction tolerance.The plot shows the comparison of the three network architectures, namely RibCage (full line), Siamese (dotted line) and Naive (dashed line) networks. All configurations were tested using two segmentation methods: KTH-SE(1) ( blue and green)  and CVUT-CZ (orange and red), on two datasets: Fluo-N2DH-GOWT1 (blue and orange) and Fluo-N2DL-HeLa (green and red)}\label{fig:hit_rate_arch}

\end{figure}

We compare three alternative architectures for the QANet, the proposed RibCage, the Siamese and the naive feed forward networks. In our
experiments, all the layers have the same number of features and end
with three FC layers.
The networks differ in the first four layers:\\
\textbf{The RibCage network} is composed of four rib-cage blocks followed by three FC layers as described as  in Section~\ref{subsec:RibCage Network Architecture}.\\
\textbf{The Siamese network} is comprised of two independent streams of four convolutional layers, one getting $I$ as input and the other $\Gamma_{\scriptsize{\mbox{E}}}$. The outputs of the last convolutional layers are concatenated and fed into the FC layers.\\
\textbf{The naive network} gets a single input, the concatenation on the channel axis of the image with evaluated segmentation image, $I$ and $\Gamma_{\scriptsize{\mbox{E}}}$. It is comprised of four convolutional layers followed by the FC layers.\\
 We tested the tree alternatives on the outputs of the CVUT-CZ  and
 the KTH-SE(1) methods applied to two datasets: Fluo-N2DH-GOWT1 and Fluo-N2DL-HeLa as described in Section~\ref{sub:TestData}.
The Hit Rate curves and the corresponding AUC scores (Section~\ref{subsec:QANetEvaluation}) for the four combinations of method-dataset are shown in Figure~\ref{fig:hit_rate_arch} and the first three rows of Table~\ref{tab:NetArch}, respectively.
The RibCage's AUC results outperform the other architectures for all method-dataset combinations, indicating that, on average, for a fixed prediction tolerance, a better Hit Rate is obtained. It is interesting to note that the Siamese architecture is consistently inferior, even with respect to the naive architecture, possibly since the low level features are of high significance.
\begin{table*}
\begin{center}
\begin{tabular}{c|c|c|c|c|c}
\hline
Architecture& Segmentation &\multicolumn{2}{c|}{KTH-SE(1)}&\multicolumn{2}{c}{CVUT-CZ}\\
~&Representation&~N2DH-GOWT1&~N2DL-HeLa&~N2DH-GOWT1&~N2DL-HeLa\\
\hline
Naive Network& Trinary &0.890&0.934&0.849&0.914\\
Siamese Network& Trinary &0.819&0.745&0.804&0.727\\
RibCage Network& Trinary &\textbf{0.904}&\textbf{0.947}&\textbf{0.912}&\textbf{0.944}\\
RibCage Network& Binary &0.902&0.933&0.908&0.917\\

\hline
\end{tabular}
\end{center}
\caption{The AUC scores for evaluating the segmentation predictions of
  KTH-SE(1) and CVUT-CZ methods on N2DH-GOWT1 and N2DL-HeLa datasets. 
  The table compares three network architecture alternatives: RibCage Network (with binary or trinary segmentation input), Siamese Network and Naive Network. The RibCage Network with trinary segmentation inputs is consistently better than the two alternatives}\label{tab:NetArch}
\end{table*}
\begin{figure}[t!]
\begin{center}

\includegraphics[trim={0bp 0bp 0bp 0bp},clip,
width=\fig_width \columnwidth]{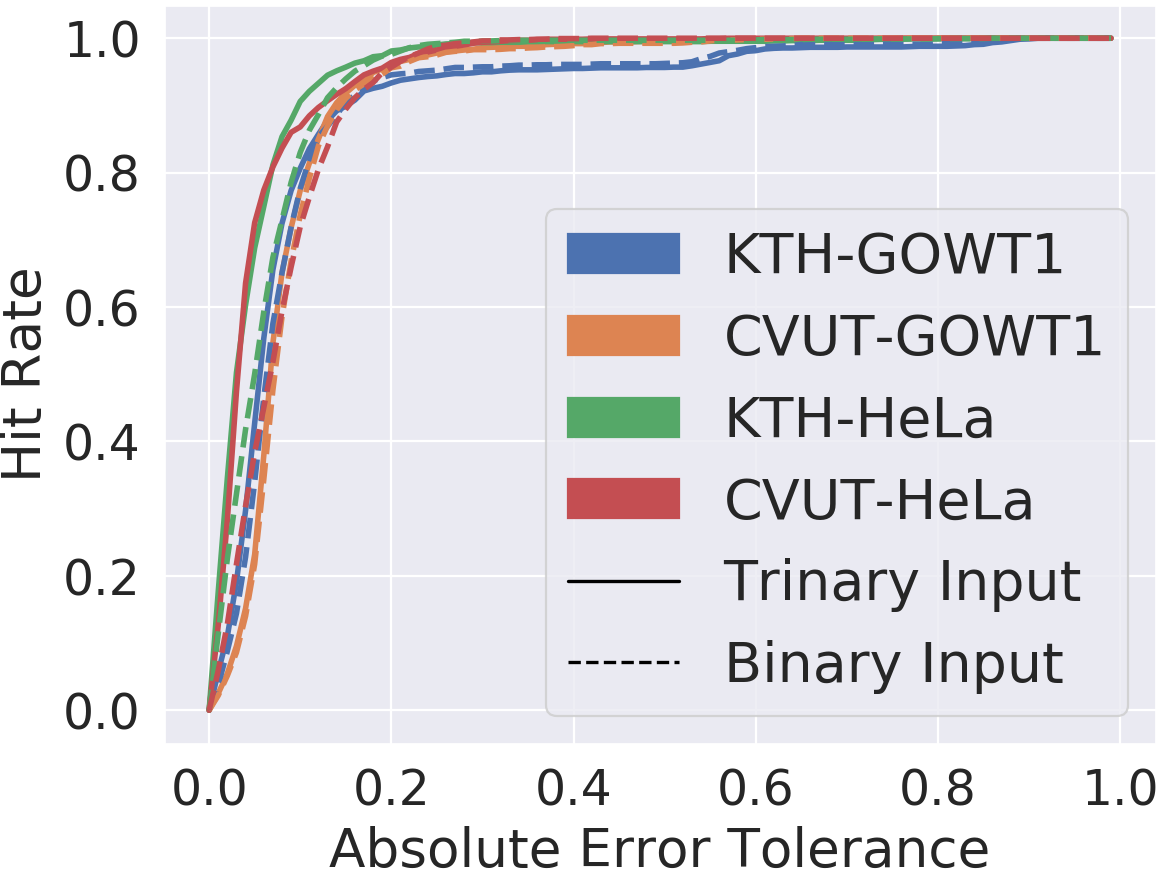} 
\end{center}
\caption{The Hit Rate curves as a function of SEG measure prediction tolerance for the RibCage Network with either binary (dashed line) or trinary (full line) segmentation input.  All configurations were tested using two segmentation methods: KTH-SE(1) (blue and green)  and CVUT-CZ (orange and red), on two datasets: Fluo-N2DH-GOWT1 (blue and orange) and Fluo-N2DL-HeLa (green and red)}\label{fig:hit_rate_TriVsBi}

\end{figure}
\subsubsection{Binary vs Trinary Segmentation Comparison}\label{subsec:2vs3Class}
While conventional segmentation methods provide binary (foreground-background) outputs, the QANet receives trinary (foreground-background-boundary) segmentations. To test which of the representations, the binary or trinary, provide better QA results, we also trained the QANet with binary input representation. The last two rows of Table~\ref{tab:NetArch} present the AUC results for the four method-dataset combinations presented in Section~\ref{sub:TestData}. The corresponding Hit Rate curves are shown in Figure~\ref{fig:hit_rate_TriVsBi}. The comparison shows that trinary segmentation is preferable. This implies that the match (or mismatch) between the boundary class and the actual instance boundaries facilitate the evaluation of the image-segmentation correspondence. 

\subsection{Cell Segmentation Benchmark Leader-board Prediction}\label{subsec:CSB Leaderboard}

\begin{table*}
\begin{center}
\begin{tabular}{c||c||c|c|c}

Evaluated Method & SEG Score & QANet Score & \multicolumn{2}{c|}{Cross-method Score \& Surrogate GT}\\
\hline
BGU-IL(3) & 0.811 & \textbf{0.808} (-0.37\%) & 0.767 (-5.42\%)& KTH-SE(1)\\
CVUT-CZ & 0.807 & \textbf{0.813} (+0.74\%) & 0.769 (-4.70\%)& KTH-SE(1)\\
KTH-SE(1) & 0.791 &\textbf{ 0.799} (+1.01\%) & 0.772 (-2.40\%)& CVUT-CZ\\
\end{tabular}
\end{center}
\caption{Predicted SEG score results for the Cell Segmentation Benchmark SIM+ dataset 
for three leading segmentation methods: 
BGU-IL(3), CVUT-CZ and KTH-SE(1) of the benchmark. 
Prediction was done for the \textbf{test data}, where the GT segmentations of this data are unknown to us. The true SEG scores are according to the Benchmark web-page. The table presents estimated QANet and the cross-method evaluation with respect to a surrogate GT. In brackets are the relative errors from the true SEG score }\label{tab:CTCPred}

\end{table*}
The prediction capabilities of the QANet were tested on the outputs of
BGU-IL(3), CVUT-CZ and KTH-SE(1). Each method was applied to the Fluo-N2DH-SIM+ \textbf{test set}. We note that the GT annotations for the test set are unavailable,
however the final scores, as validated by the benchmark organizers,
are published on the benchmark website. We then measured the mean
output of the QANet and the cross method evaluation
score. Table~\ref{tab:CTCPred} shows the true and predicted SEG
scores.
\subsubsection{Cross-method Evaluation and Surrogate GT}\label{subsec:SurrogateGT}
An alternative approach to the QANet could be the cross evaluation between multiple segmentation methods. For example, given two segmentation methods, one could act as a surrogate GT segmentation for the other, and vice versa. While this approach is valid, we show in the last column of Table~\ref{tab:CTCPred} that it is significantly less accurate than the QANet. Our assumption is that regardless of the method - a
classical or a machine learning one - segmentation processes are guided by similar principles, therefore, it is not unlikely that different segmentation methods will fail on similar examples and thus fail to evaluate each other.  
\subsection{Leaf Segmentation Challenge}\label{subsec:LSC}
We further evaluate the QANet on the Leaf Segmentation Challenge. The target quality measure was set to be the BestDice measure as is used in the challenge.
\subsubsection{The Best Dice Measure}\label{BestDice}
The Best Dice (BD) measure measure, introduced in~\cite{MinerviniPRL2015} for leaf segmentation evaluation, is an extension of the Dice measure for multiple instances in the Leaf Segmentation Challenge \cite{bell2016aberystwyth,MinerviniPRL2015,scharr2014annotated}.
For each object in the evaluated segmentation, the measure calculates the highest Dice score among all the objects in the GT segmentation and averages over all objects in the evaluated segmentation.
Let $K$ and $K'$ be the number of individual leafs in the GT and
evaluated segmentation, $\Gamma_{\scriptsize{\mbox{GT}}}$ and $\Gamma_{\scriptsize{\mbox{E}}}$ respectively.
Let $c\in\Gamma_{\scriptsize{\mbox{GT}}}$ and $c'\in\Gamma_{\scriptsize{\mbox{E}}}$ define corresponding components in
the evaluated and GT segmentations, respectively. 
The BD measure is defined as the  formulated as follows:
\begin{equation}
Q(\Gamma_{\scriptsize{\mbox{GT}}},\Gamma_{\scriptsize{\mbox{E}}}) =\frac{1}{K'}\sum_{c'\in\Gamma_{\scriptsize{\mbox{E}}}}\max_{c\in\Gamma_{\scriptsize{\mbox{GT}}}}\frac{2|c\cap c'|}{|c|+|c'|}
\end{equation}

\subsubsection{Leaf Segmentation Training Data}\label{sub: LSC Training Data}
\begin{figure*}[t!]
\begin{center}

\setlength{\tabcolsep}{0mm}
\renewcommand{\arraystretch}{0.2}
\includegraphics[trim={0bp 0bp 0bp 0bp},clip, width=\fig_width \columnwidth]{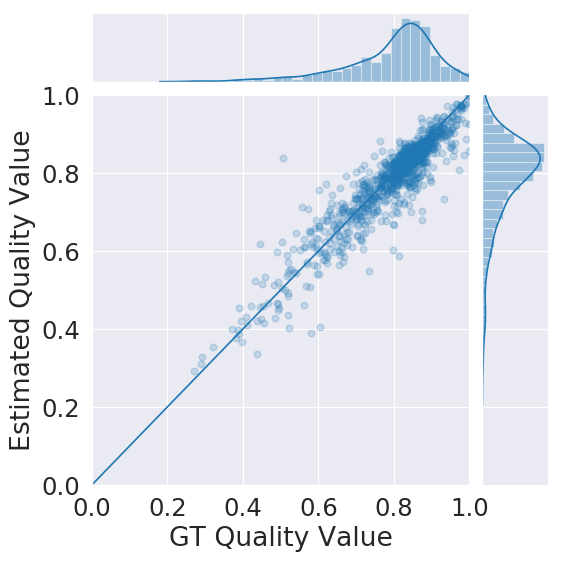}
~~~~~~~~
\includegraphics[trim={0bp 0bp 0bp 0bp},clip, width=\fig_width \columnwidth]{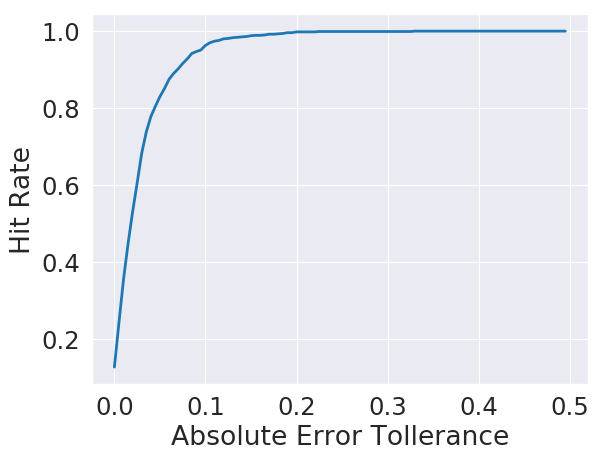}

\end{center}
\caption{\textbf{Leaf Segmentation Validation Set:} The left image shows the scatter plot of the GT quality values with respect to the estimated quality value for the leaf segmentation validation images. The horizontal axis is the GT BD of the instance and the vertical axis show the QANet output. The diagonal line represents the optimal, desired, output. On the right is a Hit Rate curve as a function of BD prediction tolerance.}\label{fig:LSC_Val}
\end{figure*}
Training the QANet, in both training and test phases, is composed of image-segmentation pairs. In the training phase we use simulated images such that the corresponding GT segmentations are known. To synthesize  imperfect segmentation proposals we deform the GT segmentations such that the true value of the quality measure, $Q$, can be calculated.
The dataset was split into a training and validation set, 70\%-30\% respectively.
The QANet was trained to regress the BD measure (described in Section~\ref{BestDice}). The scatter plot and the Hit Rate curve on the validation set are presented in Figure~\ref{fig:LSC_Val}\\
\textbf{Raw Images and GT Segmentations: }
The training images and GT segmentations were downloaded from the Leaf Segmentation Challenge web site. The images are RGB images of plants of varying sizes, where each plant has a different number of leaves.\\
\textbf{Synthesizing Imperfect Segmentations:}
If $OP$ stands for either of the morphological operations, a random integer is sampled form a integer uniform distribution: $\sigma_{\scriptsize{\mbox{MO}}}\sim U(1,6)$ which determines the size of the MO kernel. The rest of the parameters are set identically as in Section~\ref{sub:Training Data}.

\subsubsection{Leaf Segmentation Challenge Prediction}\label{subsec:LSC Real Data}
\begin{table*}
\begin{center}
\begin{tabular}{c||c||c|c|}

Evaluated Method & BD Score & QANet Score & Relative Error \\
\hline
Kuznichov et al. & 0.884 & \textbf{0.879} & -0.4\%\\
Kuznichov et al. Corrupted - Three Repetitions & 0.811$\pm$0.001 & \textbf{0.788$\pm$0.002 } &-2.7\%$\pm$0.19\% \\
\end{tabular}
\end{center}
\caption{Predicted Best Dice score results for the Leaf Segmentation Challenge dataset 
for the Kuznichov et. al. method. Due to the high accuracy of the method, we also present the results on a synthetically corrupted version of the method. The mean and standard deviation of the results for the corrupted method are shown in the second row of the table.
Prediction was done for the \textbf{test data}, where the GT segmentations of this data are unknown to us. The true Best Dice scores are according to the Challenge web-page.}\label{tab:LSCPred}
\end{table*}
The prediction capabilities of the QANet were tested on the outputs of the leaf segmentation method proposed by Kuznichov et al. \cite{Kuznichov2019}. 
We note that the GT annotations for the test set are unavailable,
however the final scores, as validated by the challenge organizers, are available for download once the results are submitted. 
We measured the mean output of the QANet and the cross method evaluation
score. The first row Table~\ref{tab:LSCPred} shows the true and predicted BD
scores.\\
\textbf{Corrupted Segmentations: }\label{subsubsec:LSC corrupted}
Due to the high accuracy of the Kuznichov et al. method, the results do not represent the capabilities of the QANet to estimate the full range of values of the quality measure. The following test is designed to achieve a wide range of segmentation quality by corrupting the original results obtained from the Kuznichov et al. method. For each image a random deformation was applied as described in \ref{subsec:Synthesized Segmentations}. The results were then submitted to the Leaf Segmentation Challenge to obtain the true score for the corrupted segmentation. The process was repeated three times and the results are presented in the last three rows of Table~\ref{tab:LSCPred}
\section{Discussion and Future Work}\label{sec:Summary}
In this paper, we introduced the Quality Assurance Deep Neural Network - QANet - a method for estimating the quality of instance segmentation at the single image level without the need for a human in the loop.

The QANet does not in itself produce a segmentation of an image, but rather predicts a quality measure of a proposed segmentation as if the GT annotation were given. Paraphrasing the British statesman Benjamin Disraeli  ``\ldots it is easier to be critical than to be correct''. 

The QANet solves a regression problem getting as input an image with its corresponding evaluated segmentation and outputting a scalar representing the estimated quality measure.
This is accomplished by a RibCage architecture~\cite{arbelleISBI2018} which inherently compares multi-level, multi-scale features of the two inputs. 

During the training phase we cover the entire range of the target quality measure by using synthesized segmentations generated by sampling random deformations of the GT segmentation. Alternatively, if the scores of the evaluated segmentations were somehow available, the actual GT segmentations are not required.

The results, based on the publicly available Cell Segmentation Benchmark and the Leaf Segmentation Challenge datasets, presented in Section~\ref{sec:Experiments}
show the QANet's ability to learn different definitions of quality measures such as the SEG measure (Sec~ \ref{subsec:The SEG Measure}) and the Best Dice measure (Sec~\ref{BestDice}). The QANet is also shown to generalize to different datasets and
segmentation methods, while being trained only on simulated data. Specifically in the case of the Cell Segmentation Benchmark, the QANet predicted the average SEG score as calculated by the benchmark organizers with maximum relative error of 1\%. These results outperform the possible alternative of the surrogate GT, as shown in Table~\ref{tab:CTCPred}. In the case of the Leaf Segmentation Challenge, the QANet also accurately predicted the mean Best Dice measure and achieved a maximum relative error of 2.7\%.

The main contribution of the QANet is providing an objective way to assess segmentations of any source. It has practical implications for the endpoint users of the segmentation methods. Moreover, the QANet can be used to alleviate training of segmentation methods either as a ranking function for an active learning frameworks or as a direct loss function for unsupervised training.\\
The code is freely available at: TBD.



\section*{Acknowledgment}
This study was partially supported by the Negev scholarship at Ben-Gurion University (A.A.); The Kreitman School of Advanced Graduate Studies (A.A) ; The Israel Ministry of Science, Technology and Space (MOST 63551 T.R.R.)

\bibliographystyle{IEEEtran}

\bibliography{IEEEabrv,DeepSeg}
\label{Bios}
\begin{IEEEbiography}[{\includegraphics[width=1in,height=1.25in,clip,keepaspectratio]{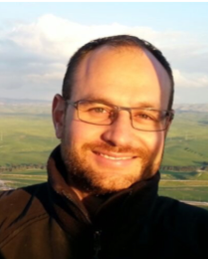}}]{Assaf Arbelle}
is a Ph.D. student at the School of Electrical and Computer Engineering of Ben-Gurion University of the Negev. His research focuses on the development of computational tools for multiple object segmentation and tracking in image sequences. He holds a B.Sc. and an M.Sc. in Electrical and Computer Engineering from Ben-Gurion University of the Negev.  
\end{IEEEbiography}
\begin{IEEEbiography}[{\includegraphics[width=1in,height=1.25in,clip,keepaspectratio]{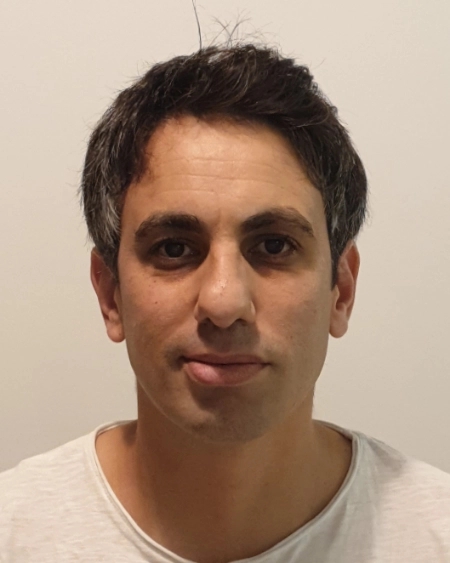}}]{Eliav Elul}
is an M.Sc. student at the School of Electrical and Computer Engineering of Ben-Gurion University of the Negev. His research interest is in the fields of computer vision and machine learning, with a focus on object segmentation and classification of biomedical images. He holds a B.Sc. in Electrical and Computer Engineering from Ben-Gurion University of the Negev.
\end{IEEEbiography}
\begin{IEEEbiography}[{\includegraphics[width=1in,height=1.25in,clip,keepaspectratio]{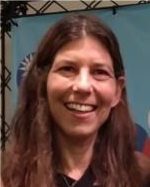}}]{Tammy Riklin~Raviv}
is a Senior Lecturer in the School of Electrical and Computer Engineering of Ben-Gurion University of the Negev.  She holds a B.Sc. in Physics and an M.Sc. in Computer Science from the Hebrew University of Jerusalem. She received her Ph.D. from the School of Electrical Engineering of Tel-Aviv University. During 2010-2012 she was a research fellow at Harvard Medical School and the Broad Institute. Prior to this (2008-2010) she was a post-doctorate associate at the Computer Science and Artificial Intelligence Laboratory (CSAIL) at the Massachusetts Institute of Technology (MIT).
\end{IEEEbiography}
\end{document}